\documentclass[journal]{IEEEtran}
\usepackage{graphicx}
\usepackage{changes}
\usepackage{amssymb}
\usepackage{graphicx}
\usepackage{amsmath,amssymb,latexsym,float,epsfig, subfigure}
\usepackage{mathtools, bbm}
\usepackage{amsmath} 
\usepackage{amssymb}  
\usepackage{lipsum}
\usepackage[export]{adjustbox}
\usepackage[normalem]{ulem} 
\usepackage{wrapfig}
\usepackage{multirow}
\usepackage{balance}
\usepackage{color}
\usepackage{url}
\usepackage{tablefootnote}
\usepackage{microtype}
\usepackage{algorithm, algorithmic}
\usepackage{breqn}
\usepackage[bottom]{footmisc}

\newcommand{\argmax}{\arg\!\max}
\newcommand{\norm}[1]{\left\lVert#1\right\rVert}

\definechangesauthor{de}

\ifCLASSINFOpdf
\else
\fi

\begin{document}

\title{Active Intent Disambiguation for \\ Shared Control Robots}

\author{Deepak~E.~Gopinath and Brenna~D.~Argall
\thanks{Deepak E. Gopinath is with the Department of Mechanical Engineering, Northwestern University, Evanston, IL and the Shirley Ryan AbilityLab, Chicago, IL.} %
\thanks{Brenna D. Argall is with the Departments of Mechanical Engineering, Computer Science and Physical Medicine and Rehabilitation, Northwestern University, Evanston, IL and the Shirley Ryan AbilityLab, Chicago, IL.}%
\thanks{Manuscript received December 2, 2018. Revised on April 10, 2020. This is the author's version of an article that has been published in this journal. Changes were made to this version by the publisher prior to publication. Final version of record is available at http://dx.doi.org/10.1109/TNSRE.2020.2987878}}
%
%

\markboth{IEEE Transactions on Neural Systems and Rehabilitation Engineering, April 2020.}
{\MakeLowercase{\textit{et al.}}: Bare Demo of IEEEtran.cls for IEEE Journals}
%

\IEEEpubid{~\copyright~2020 IEEE. Personal use is permitted. For any other purposes, permission must be obtained from the IEEE by emailing pubs-permissions@ieee.org.}

\maketitle
\begin{abstract}
Assistive shared-control robots have the potential to transform the lives of millions of people afflicted with severe motor impairments. The usefulness of shared-control robots typically relies on the underlying autonomy's ability to infer the user's needs and intentions, and the ability to do so \textit{unambiguously} is often a limiting factor for providing appropriate assistance confidently and accurately. The contributions of this paper are four-fold. First, we introduce the idea of intent disambiguation via control mode selection, and present a mathematical formalism for the same. Second, we develop a control mode selection algorithm which selects the control mode in which the user-initiated motion helps the autonomy to \textit{maximally disambiguate} user intent. Third, we present a pilot study with eight subjects to evaluate the efficacy of the disambiguation algorithm. Our results suggest that the disambiguation system (a) helps to significantly reduce task effort, as measured by number of button presses, and (b) is of greater utility for more limited control interfaces and more complex tasks. We also observe that (c) subjects demonstrated a wide range of disambiguation request behaviors, with the common thread of concentrating requests early in the execution. As our last contribution, we introduce a novel field-theoretic approach to \textit{intent inference} inspired by dynamic field theory that works in tandem with the disambiguation scheme.
\end{abstract}


%
\IEEEpeerreviewmaketitle

\section{Introduction}
\IEEEPARstart{A}{ssistive} and rehabilitation machines---such as robotic arms and smart wheelchairs---have the potential to transform the lives of millions of people with severe motor impairments~\cite{laplante1992assistive}. With rapid technological advancements in the domain of robotics these machines have become more capable and complex, and with this complexity the control of these machines has become a greater challenge. 

The standard usage of these assistive machines relies on manual teleoperation typically enacted through a control interface such as a joystick. However, the greater the motor impairment of the user, the more limited are the interfaces available for them to use. These interfaces (for example, sip-and-puffs and switch-based head arrays) are low-dimensional, at times discrete, and can typically only operate in subsets of the entire control space (referred to as \textit{control modes}). The dimensionality mismatch between the interface and the robot's controllable degrees-of-freedom (DoF) necessitates the user to switch between control modes during teleoperation and has been shown to add to the cognitive and physical burden of operation and to affect task performance negatively~\cite{herlant2016assistive}.

The introduction of robotics \textit{autonomy} to these assistive machines can alleviate some of the above-mentioned issues. More specifically, with \textit{shared} autonomy the task responsibility is shared between the user and the underlying autonomy. However, for autonomy to be effective, it needs to have a good understanding of the user's needs and intentions. That is, \textit{intent inference} is critical to ensure appropriate assistance. 

In this work, we consider use-case scenarios in which the autonomy's inference of user intent is exclusively informed by the human's control commands issued via the control interface. As an example, in the domain of assistive robotic manipulation, these control commands are typically mapped to the end-effector (or joint) velocities and result in robot motion. Motion carries information regarding underlying intent. However, intent inference becomes particularly challenging when the user input is low-dimensional and sparse---as is the case with the more limited interfaces available to those with severe motor impairments---because the robot motion will likely be more discontinuous and jagged, and carries less direct information regarding the underlying human intent. While to augment the human-robot system with high-fidelity sensors could enhance the autonomy's intent inference capabilities, for reasons of user adoption and cost, within the assistive domain we intentionally design our assistance add-ons to be as invisible and close to the manual system as possible. The need for intent \textit{disambiguation} arises as the autonomy needs to reason about all possible goals before issuing appropriate assistance commands. 
\IEEEpubidadjcol
Our key insight in this work is that user control commands issued in certain control modes are \textit{more intent expressive} than others and therefore may help the autonomy to improve inference accuracy. 
More specifically, in this work we investigate how the selection of a subset of the operational control dimensions or modes improves the intent inference and disambiguation capabilities of the robot. The idea is that earlier and more accurate inference enables the autonomy to assist the human earlier and more effectively, and thereby improve overall task performance. This is important especially in the domain of assistive robotics, wherein the purpose of the autonomy is to bridge gaps in control proficiency that result from human impairments and limited control interfaces. The main contributions of this work are as follows:
\begin{enumerate}
	\item We introduce the idea of intent disambiguation via control mode selection, and present a mathematical formalism that frames disambiguation as the problem of selecting the control mode able to maximally disambiguate between goals.
	\item We develop a control mode selection algorithm which selects the control mode \textit{for} the user, in which the user-initiated motion will help the autonomy to \textit{maximally disambiguate} intent by eliciting more \textit{intent expressive} control commands from the human.
	\item We present results from a pilot study conducted to evaluate the efficacy of the disambiguation algorithm.
	\item We propose a novel field-theoretic approach to intent inference based on ideas from \textit{dynamic field theory} in which the time evolution of the probability distribution over goals is specified as a continuous-time constrained dynamical system that obeys the principle of maximum entropy in the absence of user control commands.
\end{enumerate}

In Section~\ref{sec:related-work} we present an overview of relevant research in the areas of shared autonomy in assistive robotics, intent inference, and synergies in human-robot interaction. Section~\ref{sec:ma} presents our mathematical formalism for intent disambiguation. The disambiguation and intent inference algorithms are outlined in Sections~\ref{sec:disamb_alg} and \ref{sec:inference} respectively. The study design and experimental methods are discussed in Section~\ref{sec:study_methods} followed by results in Section~\ref{sec:results}. Discussion and conclusions are presented respectively in Sections~\ref{sec:discussion} and~\ref{sec:conclusions}.

\section{Related Work}\label{sec:related-work}
This section provides an overview of related research in the domains of shared autonomy in assistive robotics, intent inference in human-robot interaction, and synergies in human-robot interaction. 

Shared control in assistive systems aims to reduce the user's cognitive and physical burden during task execution, typically without having the user relinquish complete control~\cite{philips2007adaptive, demeester2008user, gopinath2017human, muelling2017autonomy}.
In order to offset the drop in task performance due to shifting focus from the task at hand to switching between different control modes, various mode switch assistance paradigms have been proposed. A simple time-optimal mode switching scheme has shown to improve task performance~\cite{herlant2016assistive, pilarski2012dynamic}. 

Shared-control systems often require a good estimate of the human's intent---for example, their intended reaching target in a manipulation task or a target goal location in a navigation task~\cite{liu2016goal}. Intent can be explicitly communicated by the user~\cite{choi2008laser} via various modalities such as laser pointers, click interfaces, and in some cases natural language~\cite{cantrell2010robust}. Intent can also be inferred from the user's control signals and other environmental cues using various algorithms~\cite{jain2018recursive}. Within the context of shared autonomy a Bayesian scheme for user intent prediction models the user within the Markov Decision Process framework~\cite{dragan2012formalizing, javdani2017shared, admoni2016predicting} and is typically assumed to be noisily optimizing some cost function for their intended goal. In low-dimensional spaces, this cost function can be learned from expert demonstrations using Inverse Reinforcement Learning~\cite{ziebart2008maximum}. 

For high-dimensional spaces, such as that of robotic manipulation, learning cost functions that generalize well over the entire space requires large number of samples. In such cases, heuristic cost functions, such as sum of squared velocities along a trajectory, have been found to be useful for goal prediction~\cite{dragan2013policy}. Simple heuristic approaches can also be used to find direct mappings from instantaneous cues and the underlying human intention. Heuristic approaches can incorporate domain-specific knowledge easily and are computationally inexpensive, though the trade-off for this simplicity is not being sophisticated enough to incorporate histories of states and actions, making them less robust to external noise. Instantaneous confidence functions for estimating the intended reaching target are employed with success on multiple robotic manipulation systems~\cite{gopinath2017human, abbott2007haptic}. In our work we develop an inference algorithm that updates the belief over goals using ideas from dynamic field theory in which the histories of states and actions are incorporated using a single time-scale parameter and robustness to noise is ensured via recurrent self-interactions that stabilize the dynamical system.

From the robot's perspective, the core idea behind our intent disambiguation approach is one of \textit{``Help Me, Help You''}---that is, if the user can help the autonomy with disambiguation via more intent-expressive actions, then the autonomy in turn can provide appropriate task assistance more swiftly and  accurately. In human-robot interaction, the legibility and predictability of robot motion \textit{to} the human is investigated~\cite{dragan2013legibility} with various techniques to generate legible robot motion proposed~\cite{holladay2014legible}. Our work instead investigates the idea of \textit{inverse legibility}~\cite{gopinath2017mode} in which the assistance scheme is intended to bring out more legible intent-expressive control commands \textit{from} the human. 

\section{Mathematical Formalism for Intent Disambiguation}\label{sec:ma}
This section frames intent disambiguation as a problem of determining the control mode able to maximally disambiguate between goals.

\subsection{Notation}
Let $\mathcal{G}$ be the set of all candidate goals with $n_g = |\mathcal{G}|$ and $g^i$ refer to the $i^{th}$ goal with $i \in [1,2,\dots,n_g]$. A \textit{goal} in this context represents the human's underlying intent. Specifically, in assistive robotic manipulation, as the robotic arm first must reach toward and grasp discrete objects in the environment, intent inference is the estimation of the belief over all possible discrete goals (objects) in the environment. At any time $t$, the autonomy maintains a probability distribution over goals denoted by $\boldsymbol{p}(t)$ such that $\boldsymbol{p}(t) = [p^1(t), p^2(t),\dots, p^{n_g}(t)]^{T}$ where $p^i(t)$ denotes the probability associated with goal $g^i$. The probability $p^i(t)$ represents the robot's belief that goal $g^i$ is the human's intended goal.

Let $\mathcal{K}$ be the set of all controllable dimensions of the robot and $k^i$ represent the $i^{th}$ control dimension where $i \in [1,2,\dots,n_k]$ with $n_k = |\mathcal{K}|$. The limitations of the control interface necessitate $\mathcal{K}$ to be partitioned into control modes. Let $\mathcal{M}$ be the set of all control modes with $n_m = \vert\mathcal{M}\vert$. Additionally, let $m^i$ refer to the $i^{th}$ control mode where $i \in [1,2,\dots,n_m]$. Each control mode $m^i$ is a subset of $\mathcal{K}$ such that $\bigcup\limits_{i=1}^{n_m} m^i = \mathcal{K}$.\footnote{Note that a dimension $k \in \mathcal{K}$ can be an element of multiple control modes.}

In this work, we assume a kinematic model for the robot and the kinematic state (the robot's end-effector pose) at any time $t$ is denoted as $\boldsymbol{x}_r(t) \in \mathbb{R}^3 \times \mathbb{S}^3$ and consists of a position and orientation component, where $\mathbb{S}^3$ is the space of all unit quaternions. The pose for goal $g \in \mathcal{G}$ is denoted as $\boldsymbol{x}_g \in \mathbb{R}^3 \times \mathbb{S}^3$. The control command issued by the human via the control interface is denoted as $\boldsymbol{u}_h$ and is mapped to the Cartesian velocity of the robot's end-effector. For a 6-DoF robotic arm, $\boldsymbol{u}_h \in \mathbb{R}^6$. The autonomous control policy generates an autonomy control command which is denoted as $\boldsymbol{u}_a \in \mathbb{R}^6$. The control command issued to the robot, which is a synthesis of $\boldsymbol{u}_h$ and $\boldsymbol{u}_a$ is denoted as $\boldsymbol{u} \in \mathbb{R}^6$. The control command that corresponds to a unit velocity vector along control dimension $k$ is denoted as $\boldsymbol{e}^k$.

\subsection{Disambiguation Metric}\label{ssec:disamb}
The disambiguation metrics that we develop are \textit{heuristic} measures that characterize the intent disambiguation capabilities of robot motion along particular control dimensions within particular control modes. Specifically, we define disambiguation metric $D_k \in \mathbb{R} \;\forall\; k \in \mathcal{K}$. We further explicitly denote disambiguation measures for both positive and negative motions along $k$ as $D_k^{+}$ and $D_k^{-}$ respectively. We also define a disambiguation metric $D_m \in \mathbb{R}$ for each control mode $m \in \mathcal{M}$.
By virtue of design, the disambiguation metric $D_m$ is a measure of how useful robot motion would be to the autonomy's ability to perform intent inference if the user were to operate the robot in control mode $m$. In this work our computation of $D_k$ depends on four features (denoted as $\Gamma_k$, $\Omega_k$, $\Lambda_k$ and $\Upsilon_k$), that capture different aspects of the \textit{shape} of a projection of the probability distribution over intent. These projections and computations are described in detail in Section~\ref{ssec:projection} and Section~\ref{ssec:components}, and as pseudocode in Algorithm~\ref{alg1}. 

\section{Disambiguation Algorithm}\label{sec:disamb_alg}
This section describes the computation of $D_k$ and $D_m$ contributed in this article, as well as our intent disambiguation algorithm.
\subsection{Forward Projection of $\boldsymbol{p}(t)$}\label{ssec:projection}
The first step in the computation of $D_k$ is a model-based forward projection of the probability distribution $\boldsymbol{p}(t)$ from the current time $t_a$ to times $t_b$ and $t_c$ (Algorithm~\ref{alg1}, line 4) where $t_a < t_b < t_c$.\footnote{\textit{UpdateIntent()} in Line 4 is implemented using Equation~\ref{eq:dft_ii} discussed in detail in Section~\ref{ssec:dft_ii}. \textit{SimulateKinematics()} assumes that the end-effector kinematics is same as that of a point-like robot. All parameters which affect the computation of $\boldsymbol{p}(t)$ are denoted as $\boldsymbol{\Theta}$.} We consider two future times in order to compute short-term ($t_b$) and long-term ($t_c$) evolutions of the probability distribution. The application of unit velocity $\boldsymbol{e}^k$ results in probability distributions $\boldsymbol{p}^+_k(t_b)$ and $\boldsymbol{p}^+_k(t_c)$, and $-\boldsymbol{e}^k$ results in $\boldsymbol{p}^-_k(t_b)$ and $\boldsymbol{p}^-_k(t_c)$, where the subscript $k$ captures the fact that the projection is the result of the application of a control command only along control dimension $k$. 

\begin{algorithm}[t]
	\caption{Intent Disambiguation}
	\label{alg1}
	\begin{algorithmic}[1]
		\REQUIRE $\boldsymbol{p}(t_a), \boldsymbol{x}_r(t_a), \Delta t, t_a < t_b < t_c, \boldsymbol{\Theta}$
		\FOR{$k=0\dots n_k$}
		\STATE Initialize $D_k = 0$, $t = t_a$, $\boldsymbol{u}_h = \boldsymbol{e}^k$
		
		\WHILE{$t \leq t_c$}
		\STATE $\boldsymbol{p}_k(t + \Delta t) \leftarrow \text{UpdateIntent}(\boldsymbol{p}_k(t), \boldsymbol{u}_h; \boldsymbol{\Theta})$
		\STATE $\boldsymbol{x}_r(t + \Delta t) \leftarrow \text{SimulateKinematics}(\boldsymbol{x}_r(t), \boldsymbol{u}_h)$
		\IF{$t = t_b$} \STATE {$Compute \;\;\Gamma_k, \Omega_k, \Lambda_k$} 
		\ENDIF
		\IF{$t = t_c$} \STATE{$Compute \;\;\Upsilon_k$} \ENDIF
		\STATE $t \leftarrow t + \Delta t$
		\ENDWHILE
		\STATE $Compute \;\;D_k$
		\ENDFOR
	\end{algorithmic}
\end{algorithm}
\subsection{Features of $D_k$}\label{ssec:components}
To compute our control dimension disambiguation metric, we design four features that encode different aspects of the \textit{shape} of the probability distribution as it evolves under motion in a specific control dimension $k$. For each control dimension $k$, each of the four features is computed for projections along both positive and negative directions independently. The four features are computed in lines 7 and 10 in Algorithm~\ref{alg1}.

1) \textit{Maximum:} The maximum of the projected probability distribution $\boldsymbol{p}_k(t_b)$ is a good measure of the robot's \textit{overall certainty} in accurately predicting human intent. We define the distribution maximum as
\begin{equation}
\Gamma_k = \max\limits_{1 \leq i \leq n_g}p^i_k(t_b)
\end{equation}
(i.e., the statistical mode of this discrete distribution).
A higher value implies that the robot has a greater confidence in its prediction of the human's intended goal.

2) \textit{Pairwise separation:} More generally, disambiguation accuracy benefits from a larger separation, $\Lambda_k$, between goal probabilities. The quantity $\Lambda_k$ is computed as the \textit{sum of the pairwise distances} between the $n_g$ probabilities.
\begin{equation}
\Lambda_k = \sum_{i=1}^{n_g}\sum_{j=i}^{n_g}\lvert p^i_k(t_b) - p^j_k(t_b)\rvert
\end{equation}
$\Lambda_k$ is particularly helpful if the difference between the largest probabilities fails to disambiguate.

3) \textit{Difference between maxima:} Disambiguation accuracy benefits from greater differences between the first and second most probable goals. This difference is denoted as
\begin{equation}
\Omega_k = \text{max}(\boldsymbol{p}_k(t_b)) - \text{max}(\boldsymbol{p}_k(t_b) \setminus \text{max}(\boldsymbol{p}_k(t_b))) 
\end{equation}
and $\Omega_k$ becomes particularly important when the distribution has multiple modes and a single measure of maximal certainty ($\Gamma_k$) alone is not sufficient for successful disambiguation. 

4) \textit{Gradients:} $\Gamma_k, \Omega_k$ and $\Lambda_k$ are local measures that encode shape characteristics of the short-term temporal projections of the probability distribution over goals. However, the quantity $\boldsymbol{p}_k(t)$ can undergo significant changes upon long-term continuation of motion along control dimension $k$. The spatial gradient of $\boldsymbol{p}_k(t)$ encodes this propensity for change and is approximated by 
\begin{equation*}
\frac{\partial\boldsymbol{p}_k(t)}{\partial x_k} \simeq \frac{\boldsymbol{p}_k(t_c) - \boldsymbol{p}_k(t_b)}{x_k(t_c) - x_k(t_b)}
\end{equation*}
where $x_k$ is the component of robot's projected displacement along control dimension $k$. The greater the difference between individual spatial gradients, the greater will the probabilities deviate from each other, thereby helping in disambiguation. In order to quantify the ``spread'' of gradients we define $\Upsilon_k$ as
\begin{equation}
\Upsilon_k = \sum_{i=1}^{n_g}\sum_{j=i}^{n_g}\Big \lvert\frac{\partial p^i_k(t)}{\partial x_k} - \frac{\partial p^j_k(t)}{\partial x_k}\Big \rvert
\end{equation}
where $\lvert\cdot\rvert$ denotes the absolute value. 

5) \textit{Computation of $D_k$ and $D_m$:} 
The individual features $\Gamma_k$, $\Omega_k$, $\Lambda_k$ and $\Upsilon_k$ are combined to compute $D_{k}$ in such a way that, by design, higher values of $D_k$ imply greater disambiguation capability for the control dimension $k$. More specifically, 
\begin{equation}\label{DK}
D_{k} = \underbrace{w\cdot(\Gamma_k\cdot \Lambda_k\cdot\Omega_k)}_{\text{short-term}} + \underbrace{(1 - w)\cdot \Upsilon_k}_{\text{long-term}}
\end{equation}
where $w$ is a task-specific weight that balances the contributions of the short-term and long-term components. In our implementation, $w$ is empirically set to $0.5$. Equation~\ref{DK} is computed twice, once in each of the positive ($\boldsymbol{e}^k$) and negative directions ($-\boldsymbol{e}^k$) along $k$, and the results ($D_k^+$ and $D_k^-$) are then summed to compute $D_k$. 

The computation of $D_k$ is performed for each control dimension $k \in \mathcal{K}$. The disambiguation metric $D_m$ for control mode $m$ then is calculated as 
\begin{equation*}\label{EQ2}
D_m = \sum_{k \in m} D_{k} \;
\end{equation*}
and the control mode with highest disambiguation capability $m^*$ is given by $m^* = \argmax_m  D_{m}$
while $k^* = \argmax_k D_k$ gives the control dimension with highest disambiguation capability $k^{*}$.
Disambiguation mode $m^{*}$ is the mode that the algorithm chooses \textit{for} the human to better estimate their intent. 
\section{Intent Inference}\label{sec:inference}
Since the disambiguation power of our algorithm is closely linked to the fidelity of the underlying intent inference mechanism,
in this section, we propose a novel intent inference scheme inspired by \textit{dynamic field theory}. By having the autonomy maintain a probability distribution over goals, we implicitly model the human as a Partially Observable Markov Decision Process (POMDP) in which all the uncertainty in the user's state is concentrated in the user's intended goal. By maintaining and updating a probability distribution over goals the autonomy can reason about the human's latent state (internal goal) during trial execution.  Inference over goal states typically is done using a recursive Bayesian belief update which determines how the distribution evolves over time. Here we introduce a novel approach to compute the time evolution of a probability distribution over goals as a dynamical system with constraints which serves as an alternative to the recursive Bayesian update scheme.

\subsection{Dynamic Field Theory}\label{ssec:dft}

In Dynamic Field Theory (DFT)~\cite{schoner2015dynamic}, variables of interest are treated as dynamical state variables. To represent the information about these variables requires two dimensions: one which specifies the value the variables can attain and the other which encodes the \textit{activation level} or the amount of information about a particular value. These \textit{activation fields} (also known as dynamic neural fields) are analogous to probability distributions defined over a random variable. 

Following Amari's formulation~\cite{amari1977dynamics} the dynamics of an activation field $\phi(x, t)$ are given by 
\begin{multline*}
\tau\frac{\partial{\phi}(x,t)}{\partial t} = \int\limits_{}^{}dx^{\prime}b(x-x^{\prime})\sigma(\phi(x^{\prime}, t)) -\phi(x,t) + h + S(x,t) 	
\end{multline*} 
where $x$ denotes the variable of interest, $t$ is time, $\tau$ is the time-scale parameter, $b(x-x^\prime)$ is the interaction kernel, $\sigma(\phi)$ is a sigmoidal nonlinear threshold function, $h$ is the constant resting level and $S(x,t)$ is the external input. The interaction kernel mediates how activations at all other field sites $x^\prime$ drive the activation level at $x$. Two types of interactions are possible: excitatory (when interaction is positive) which drives up the activation, and inhibitory (when the interaction is negative) which drives the activation down. 

Historically, dynamic neural fields were conceived to explain cortical population neuronal dynamics based on the hypothesis that the excitatory and inhibitory neural interactions between local neuronal pools form the basis of cortical information processing. 
These activation fields possess unique characteristics that make them ideal candidates for modeling the time evolution of $\boldsymbol{p}(t)$. First, a peak in the activation field can be \textit{sustained} even in the absence of external input due to the recurrent interaction terms. Second, information from the past can be \textit{preserved} over much larger time scales quite easily by tuning the time-scale parameter thereby endowing the fields with memory. Third, the activation fields are \textit{robust} to disturbance and noise in the external input~\cite{schoner2008dynamical}. We harness these characteristics to specify a model for smooth temporal evolution of $\boldsymbol{p}(t)$ in the next section. 

\subsection{Field-Theoretic Intent Inference}\label{ssec:dft_ii}
Our insight is to use the framework of dynamic neural fields to specify the time evolution of the probability distribution $\boldsymbol{p}(t)$, in which we treat the individual goal probabilities $p^i(t)$ as constrained dynamical state variables such that $p^i(t) \in [0, 1]$ and $\Sigma_{1}^{n_g}p^{i}(t) = 1$. We refer to this approach as the \textit{field-theoretic intent inference.}

The full specification of the field is given by
\begin{equation*}
\frac{\partial \boldsymbol{p}(t)}{\partial t} = \frac{1}{\tau}\bigg[\underbrace{-\boldsymbol{P}^{T}_{n_g\times n_g}\cdot\boldsymbol{p}(t)}_{\text{goal_transition_dynamics}} + \underbrace{\frac{1}{n_g}\cdot\mathbbm{1}_{n_g}}_{\text{rest state}}\bigg]
\end{equation*}
\begin{equation}\label{eq:dft_ii}
+  \underbrace{\boldsymbol{\lambda}_{n_g\times n_g}\cdot\sigma(\boldsymbol{\xi}(\boldsymbol{u}_h;\boldsymbol{\Theta}))}_{\text{excitatory + inhibitory}}
\end{equation}

where time-scale parameter $\tau$ determines the memory capacity and decay behavior, $\boldsymbol{P}_{n_g\times n_g}$ is the state transition matrix for the embedded Markov chain that models the goal transitions as jump processes, $\mathbbm{1}_{n_g}$ is a vector of dimension $n_g$ containing all ones, $\boldsymbol{u}_h$ is the human control input and $\boldsymbol{\Theta}$ represents all other task-relevant features, $\boldsymbol{\lambda}$ is the control matrix that controls the excitatory and inhibitory aspects, $\boldsymbol{\xi}$ is a function that encodes the nonlinearity through which human control commands and task features affect the time evolution, and $\sigma$ is a biased sigmoidal nonlinearity given by $\sigma(\boldsymbol{\xi}) = \frac{1}{1 + e^{-\boldsymbol{\xi}}} - 0.5$.

Our design of $\boldsymbol{\xi}$ is informed by what features of the human control input and environment effectively capture the human's underlying intent. We choose the \textit{directedness} of the robot motion towards a goal, the \textit{agreement} between the human and robot autonomy commands, and the \textit{proximity} to a goal. The \textit{directedness} component looks at the shortest straight line path towards a goal $g$, whereas the \textit{agreement} serves as an indicator of how similar (measured as a dot product) the human and autonomy signals are to each other.
 One dimension $i$ of $\boldsymbol{\xi}$ is defined as 
 \begin{align*}
     \xi^i(\boldsymbol{u}_h;\boldsymbol{\Theta}) & = \underbrace{\frac{1 + \eta}{2}}_{\text{directedness}} + \underbrace{\boldsymbol{u}_{h}^{rot}\cdot\boldsymbol{u}_{a,g^i}^{rot}}_{\text{agreement}}
     \\&+\underbrace{\text{max}\Big(0, 1-\frac{\norm{\boldsymbol{x}_{g^i} - \boldsymbol{x}_r}}{R}\Big)}_{\text{proximity}}
 \end{align*}

where  $trans$ and $rot$ refer to the translational and rotational components of a command $\boldsymbol{u}$ or position $\boldsymbol{x}$, $\eta = \frac{\boldsymbol{u}_h^{trans}\cdot(\boldsymbol{x}_{g^i} - \boldsymbol{x}_r)^{trans}}{\norm{\boldsymbol{u}_h^{trans}}\norm{(\boldsymbol{x}_{g^i} - \boldsymbol{x}_r)^{trans}}}$, $\boldsymbol{u}_{a,g^i}$ is the robot autonomy command for reaching goal $g^i$,  $R$ is the radius of the sphere beyond which the proximity component is always zero, $\norm{\cdot}$ is the Euclidean norm and $\boldsymbol{\Theta} = \{\boldsymbol{x}_r, \boldsymbol{x}_{g^i}, \boldsymbol{u}_{a, g^i}\}$
At every time-step, constraints on $p^i(t)$ are enforced such that $\boldsymbol{p}(t)$ is a valid probability distribution. 
The most confident goal $g^*$ is computed as $g^* = \argmax_i  p^i(t) \;\forall\; i \in [1,\dots,n_g]$.

\subsection{Field-Theoretic Intent Inference For Assistive Robotics}\label{ssec:dft_adv}

In our work, the autonomy's inference of user intent solely relies on user control commands. In the domain of assistive robotics, it is quite often the case that the user input is highly discontinuous (due to fatigue, motor impairments, stoppage for mode switches, \textit{et cetera}). Therefore, it is important to reason about belief over goals also in the \textit{absence} of useful information. 

According to the \textit{principle of maximum entropy}, in the absence of testable information (no control commands issued by the user and a uniform global prior), the belief should converge to a uniform distribution. In the absence of $\boldsymbol{u}_h$, using Equation~\ref{eq:dft_ii} and an appropriately chosen time-scale parameter $\tau$, $\boldsymbol{p}(t)$ converges to a uniform distribution by correctly ignoring outdated information. The rate at which the distribution decays to a uniform distribution is controlled by $\tau$.

By contrast, the standard discrete-time recursive belief update equation as implemented in~\cite{jain2018recursive} is
\begin{equation*}
p(g^t |\boldsymbol{u}_h^t) = \eta p(\boldsymbol{u}_h^t | g^t)\sum_{g^{t-1} \in \mathcal{G}}^{}p(g^t | g^{t-1})p(g^{t-1} |\boldsymbol{u}_h^{t-1})
\end{equation*}
where $\eta$ is a normalization factor, $\boldsymbol{p}(\boldsymbol{u}_h | g)$ is a likelihood function, and $p(g^t | g^{t-1})$ is the goal transition probability. 
In the recursive belief update, when $\boldsymbol{u}_h = 0$ and the likelihood function is uniform, it can be shown that the posterior distribution over goals converges to the stationary distribution of the goal transition matrix. The stationary distribution is not necessarily uniform and can introduce unwanted biases in the inference. 

Knowledge of task-level semantics can provide informative global priors that can further improve the accuracy of the inference mechanism. Our field-theoretic approach additionally can encode a task-level global prior in the `rest state' term. For example, in a pick-and-place task, the initial goal distribution could be biased towards the object that needs to be picked. 
\begin{figure}[t]
	\centering
	\includegraphics[width = 1.0\hsize]{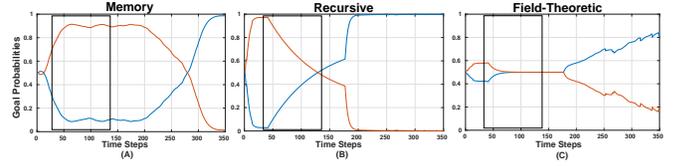}
    \caption{Inference comparison: Goal probabilities (blue and orange lines) for (A) Memory-Based Prediction, (B) Recursive Bayesian Belief Update and (C) Field-Theoretic Inference in a scene with two goals. Black rectangular boxes indicate times of zero control velocity, with varying effects on the inference schemes: (A) little change (since the cost function is purely distance-based), and convergence to (B) the stationary distribution of the goal transition matrix $\boldsymbol{P}$ and (C) the uniform distribution as dictated by the principle of maximum entropy.}
	\label{fig:dft_inference}
\end{figure}
In order to evaluate the performance of our field-theoretic inference approach a quantitative comparison to (1) memory-based prediction~\cite{dragan2013policy} and (2) recursive belief updating~\cite{jain2018recursive} was implemented using point robot simulation in $\mathbb{R}^3$. The human was modeled as issuing a control command that noisily optimizes a straight-line path towards the intended goal. Signal dropout was simulated by randomly zeroing out control commands and $\tau$ was set to be 10. Additionally, $\boldsymbol{u}_h$ was set to be zero for a randomly chosen section of each trial in order to compare the convergence behavior of different approaches. The number of goals varied between three and five. Goal transitions were randomly sampled every five to eight time steps. The average length of the simulated trajectories was 615 time steps. 500 trials were simulated. Inference accuracy was computed as the fraction of total trial time (excluding when $\boldsymbol{u}_h = 0$) for which an algorithm correctly inferred the ground truth. 

Results for field-theoretic inference outperformed memory-based prediction significantly ($87.46\%$ vs. $59.15\%$ respectively) and were comparable to recursive belief updating ($87.43\%$). Figure~\ref{fig:dft_inference} shows an illustrative example of goal inference using the various methods. One can see that when there is no control command issued, the field-theoretic approach alone converges to the maximum entropy uniform distribution.

\section{Study Methods}\label{sec:study_methods}
In this section, we describe the study methods used to evaluate the efficacy of the disambiguation system. 

\noindent{\underline{\textbf{Participants:}}} For this study eight subjects were recruited (mean age: $31 \pm 11$, 3 males and 5 females). All participants gave their informed, signed consent to participate in the experiment, which was approved by Northwestern University's Institutional Review Board.

\begin{figure}[b]
	\centering
	\includegraphics[width = 1\hsize]{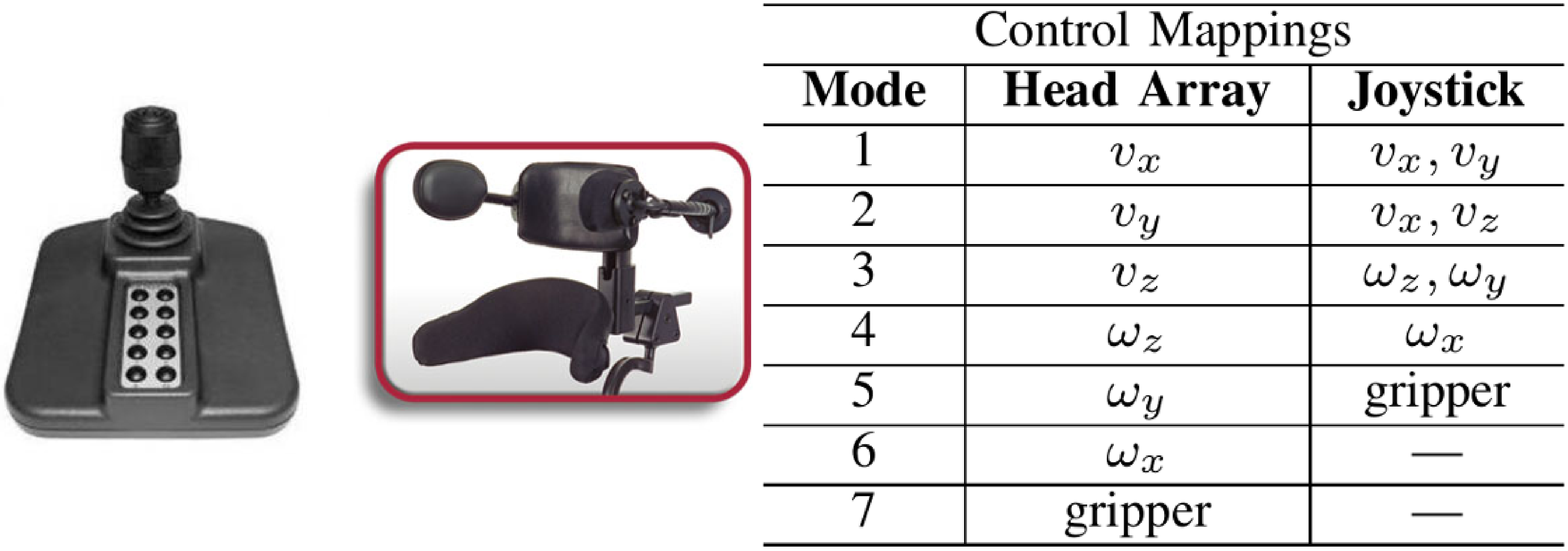}
	\caption{A 2-axis joystick (left) and switch-based head array (center) and their operational paradigms (right). $v$ and $\omega$ indicate the translational and rotational velocities of the end-effector, respectively. }
	\label{fig:interfaces}
\end{figure}

\noindent{\underline{\textbf{Hardware:}}} The experiments were performed using the MICO 6-DoF robotic arm (Kinova Robotics, Canada), specifically designed for assistive purposes. The software system was implemented using the Robot Operating System (ROS) and data analysis was performed in MATLAB. 
The subjects teleoperated the robot using two different control interfaces: a 2-axis joystick and a switch-based head array, controlling the 6D Cartesian velocity of the end-effector (Figure~\ref{fig:interfaces}). An external button was provided to request the mode switch assistance. 

In detail, the joystick generated 2D continuous control signals. Under joystick control the full control space was partitioned into five control modes that were accessed via button presses. 
The switch-based head array consisted of three switches embedded in a headrest, operated via head movements, and generated 1D discrete signals. Under head array control the full control space was partitioned into seven control modes, the back switch was used to cycle between the different control modes, and the switches to the left and right controlled the motion of the robot's end effector in the positive and negative directions along a selected control dimension.
\begin{figure}[ht!]
	\includegraphics[keepaspectratio, width = 0.49\textwidth ]{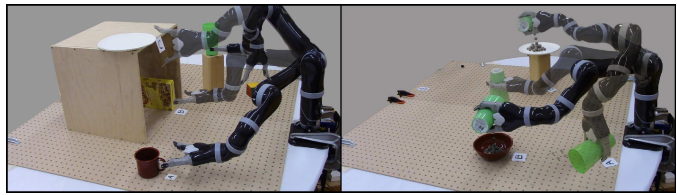}
	\caption{Study tasks performed by subjects. \textit{Left:} Single-step reaching task. \textit{Right:} Multi-step pouring task. }
	\label{fig:tasks}
\end{figure}

\noindent{\underline{\textbf{Tasks:}}} Two different categories of tasks were evaluated.

\underline{\textit{Single-step:}} The aim was to reach one of five objects on the table, each with a target orientation (Figure~\ref{fig:tasks}, Left). 

\underline{\textit{Multi-step:}} Each trial began with a full cup held by the robot gripper. The task required first that the contents of the cup be poured into one of two containers, and then that the cup be placed at one of the two specified locations and with a particular orientation (Figure~\ref{fig:tasks}, Right). 

\noindent{\underline{\textbf{Switching Paradigms:}}} Two kinds of mode switching paradigms were evaluated in the study.

\underline{\textit{Manual}}: During task execution the user performed all mode switches. 

\underline{\textit{Disambiguation}}: The user either performed a mode switch manually or requested a switch to the \textit{disambiguation} mode. The user was free to issue disambiguation requests at any time during the task execution, upon which the algorithm identified and switched the current control mode to the best disambiguation mode $m^*$ by invoking Algorithm~\ref{alg1}. To be clear, the user also was allowed to switch control modes using a manual mode switch at any time as well. The only requirement was that, the user request disambiguation at least once during the task execution. 

\noindent{\underline{\textbf{Shared Control:}}} Autonomy assistance was always active for both mode switch assistance paradigms (manual and disambiguation). We used a blending-based shared-control paradigm in which the final robot control command was a linear composition of the human control command and an autonomous control command. With blending the amount of assistance was directly proportional to the probability of the most confident goal $g^*$, and thus to the strength of the intent inference. The probability distribution over goals, $\boldsymbol{p}(t)$, was updated using Equation~\ref{eq:dft_ii} as outlined in Section~\ref{ssec:dft_ii} and the most confident goal was computed as $\argmax_i  p^i(t)$. Therefore, if intent inference improved as a result of goal disambiguation, more assistance would be provided by the autonomy.

Specifically, the autonomous control policy generated control command $\boldsymbol{u}_a \leftarrow f_{a}(\boldsymbol{x}_r)$
where $f_{a}(\cdot) \in \mathcal{F}_{a}$, and $\mathcal{F}_{a}$ was the set of all control behaviors corresponding to different tasks. This set could be derived using a variety of techniques such as \textit{Learning from Demonstrations}~\cite{argall2009survey}, motion planners~\cite{hsu2002randomized} or navigation functions~\cite{rimon1992exact}.
In our implementation, the autonomy's control command was generated using a simple potential field which is defined in all parts of the state space~\cite{khatib1986real}. Every goal $g$ was associated with a potential field $\gamma_g$ which treats $g$ as an attractor and all other goals in the scene as repellers. The autonomy command was computed as a summation of the attractor and repeller velocities and operated in the full 6D Cartesian space. 

Let $\boldsymbol{u}_{a,g}$ be the autonomy command associated with goal $g$. Under blending, the final control command $\boldsymbol{u}$ issued to the robot then was given by
\begin{equation}\label{eq:sc}
\boldsymbol{u} = \alpha\cdot \boldsymbol{u}_{a,g^*} + (1 - \alpha)\cdot \boldsymbol{u}_h
\end{equation}
where $g^*$ was the most confident goal. Similar to $\boldsymbol{u}_h$, the autonomy command $\boldsymbol{u}_{a, g^*} \in \mathbb{R}^6$ was mapped to the 6D Cartesian velocity of the end-effector. 
The blending factor $\alpha$ was a piecewise linear function of the probability $p(g^*)$ associated with $g^*$ and was given by
$$
\alpha = \left\{
\begin{array}{ll}
0 & \quad\quad~~~ p(g^*) \leq \rho_1 \\
\frac{\rho_3 (p(g^*) - \rho_1)}{\rho_2 - \rho_1}  &  \quad \text{if}\quad \rho_1 < p(g^*) \leq \rho_2  \\
\rho_3 & \quad\quad~~~  \rho_2 <  p(g^*)
\end{array}
\right.
$$
with $\rho_i \in [0, 1] \;\forall\; i \in [1,2,3]$ and $ \rho_2 > \rho_1$. 
In our implementation, we empirically set $\rho_1 = \frac{1.2}{n_g}, \rho_2 = \frac{1.4}{n_g}$ and $ \rho_3 = 0.7$.

\noindent{\underline{\textbf{Study protocol:}}} A within-subjects study was conducted using a fractional factorial design in which the manipulated variables were the tasks, control interfaces, and the switching paradigm conditions. Each subject underwent an initial training period that lasted approximately 30 minutes. 
The training period consisted of three phases and two different task configurations. The subjects used both interfaces to perform the training tasks.

\underline{\textit{Phase One}}: The subjects were asked to perform a simple reaching motion towards a single goal in the scene. This phase was intended for the subjects to get familiarized with the control interface mappings and teleoperation of the robotic arm.

\underline{\textit{Phase Two}}: Subjects were asked to perform a simple reaching motion towards a single goal in the scene in the presence of blending-based autonomous assistance.

\underline{\textit{Phase Three}}: Subjects were able to explore the disambiguation request feature during a reaching task to observe the effects of the mode switch request and subsequent change in robot assistance. Multiple objects were introduced in the scene. 
Subjects were explicitly informed that upon a disambiguation request the robot would select a control mode that would help the autonomy determine the subject's intended goal and thereby enable it to assist the user more effectively.

\begin{figure}[t!]
	\centering
	\includegraphics[keepaspectratio, width = 1.\hsize, height = 0.4\vsize, left]{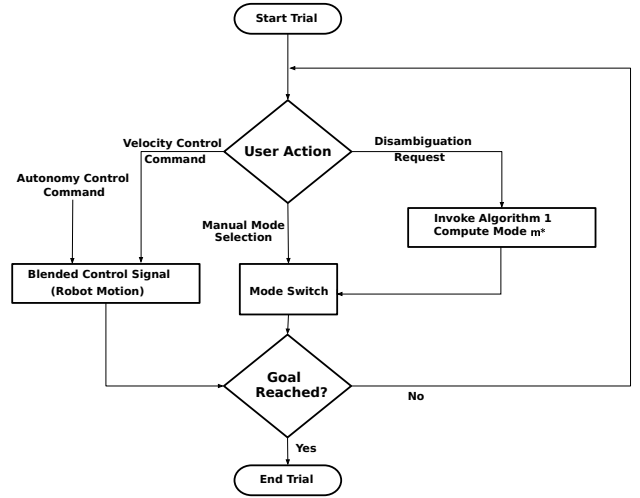}
	\caption{Flow chart depicting user action sequence during a single trial. The user could issue a (i) velocity control command, resulting in intent inference followed by generation of an autonomy signal and then blended control signal, and causes robot motion or (ii) manual mode switch or (iii) disambiguation request, both resulting in a control mode switch.}
	\label{fig:exp_block_diagram}
\end{figure}

During the testing phase, each subject performed both tasks using both interfaces under the \textit{Manual} and \textit{Disambiguation} paradigms. All trials started in a randomized initial control mode and robot position. The ordering of control interfaces and paradigms was randomized and counterbalanced across all subjects. Three trials were collected for the \textit{Manual} paradigm and five trials for the \textit{Disambiguation} paradigm. On an average, each trial lasted approximately 10-40s depending on the starting position of the robot and the specified reaching target. At the start of each trial, $\boldsymbol{p}(t)$ was initialized as $\frac{1}{n_g}\cdot\mathbbm{1}_{n_g}$. During the trial as the user teleoperated the robot, $\boldsymbol{p}(t)$ was updated according to Equation~\ref{eq:dft_ii} online at each time step. Figure~\ref{fig:exp_block_diagram} captures how a single trial unfolds in time.

\noindent{\underline{\textbf{Metrics:}}}
The objective metrics used for evaluation included the following. 
\begin{itemize}
	\item \textit{Number of mode switches}: The number of times a user switched between various control modes during task execution. This metric captures one of the main factors that contributes to the cognitive and physical effort required for task execution in assistive robotic manipulation~\cite{herlant2016assistive}.
	\item \textit{Number of disambiguation requests}: The number of times a user pressed the disambiguation request button. 
	\item \textit{Number of button presses}: The sum of \textit{Number of mode switches} and \textit{Number of disambiguation requests}.
	\item \textit{Skewness}: A higher-order moment used to quantify the asymmetry of any distribution. Used to characterize how much the temporal distribution of disambiguation requests deviates from a uniform distribution. 
	\item \textit{Task completion time}: Time taken to complete the task successfully. This metric is an indicator of how well the human and autonomy work together.
\end{itemize}

Additionally, at the end of each testing phase, subjective data was gathered via a brief questionnaire. Users were given the following statements regarding the usefulness and capability of the assistance system to rate according to their agreement on a 7-point Likert scale.
\begin{itemize}
	\item \textbf{Q1} - Control modes chosen by the system made task execution easier.
	\item \textbf{Q2} - The robot and I worked together to accomplish the task.
	\item \textbf{Q3} - I liked operating the robot in the control modes chosen by the system.
\end{itemize}
Subjects were also asked to indicate their preference in the following questions.
\begin{itemize}
	\item \textbf{Q4} - Which interface was the hardest to operate?
	\item \textbf{Q5} - For which interface was the assistance paradigm the most useful?
	\item \textbf{Q6} - Which one of the schemes do you prefer the most?
	\item \textbf{Q7} - Which one of the schemes is the most user-friendly? 
\end{itemize}

\section{Results}\label{sec:results}
\begin{figure}[b]
	\centering
	\includegraphics[keepaspectratio, width = 1\hsize ,left]{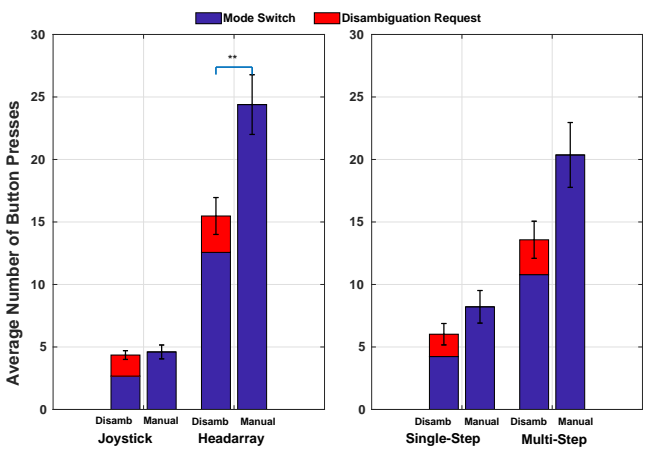}
	\caption{Average number of button presses, \textit{Disambiguation} and \textit{Manual} paradigms. \textit{Left:} Grouped by control interfaces. \textit{Right:} Grouped by tasks.}
	\label{fig:button_press}
\end{figure}
In this section we present results from our human-subject study. Our study results indicate that the disambiguation request system is of greater utility for more limited control interfaces and more complex tasks. Subjects demonstrated a wide range of disambiguation request behaviors with a common theme of relying on disambiguation assistance earlier in the trials. Furthermore, the survey results show that operating the robot in the disambiguating mode make task execution easier and that users prefer the \textit{Disambiguation} paradigm to the \textit{Manual} paradigm. Statistical significance is determined using the Wilcoxon Rank-Sum test in where (***) indicates $p < 0.001$, (**) $p < 0.01$ and (*) $p < 0.05$.
\begin{figure}[t!]
	\centering
	\includegraphics[width = 0.5\textwidth,center]{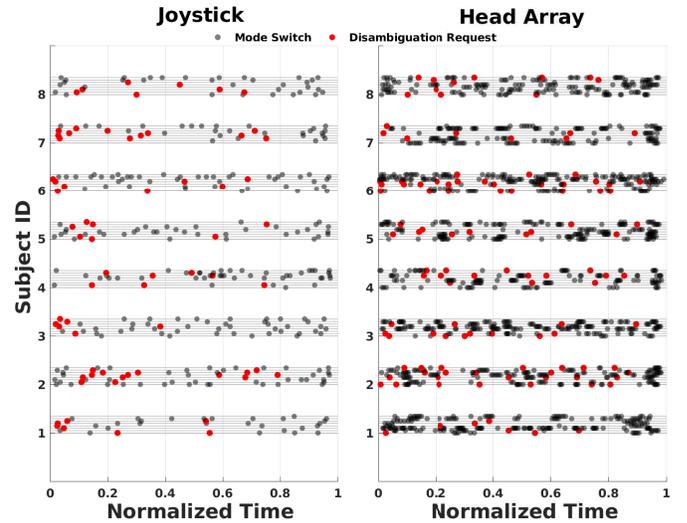}
	\caption{Temporal pattern of button presses for joystick (left) and head array (Right) during the multi-step task on a trial-by-trial basis for all subjects. For each subject, each light gray horizontal line represents a single trial. Eight trials per subject, for each interface.}
	\label{fig:ha_man_disamb}
\end{figure}

\noindent{\textbf{Impact of Disambiguation:}} A statistically significant decrease in the number of button presses is observed between the \textit{Manual} and \textit{Disambiguation} paradigms when using the head array (Figure~\ref{fig:button_press}, Left). Due to the low-dimensionality of the head array and cyclical nature of mode switching, the number of button presses required for task completion is inherently high. The disambiguation paradigm is helpful in reducing the number of button presses likely due to higher robot assistance that is present in the disambiguating control mode.
For the joystick, statistically significant differences between the two paradigms are observed for the number of manual mode switches ($p < 0.05$). However, this gain is offset by the button presses that are required to make disambiguation requests. When grouping by task, the general trend of a decrease in the number of button presses is more pronounced for the more complex multi-step task (Figure~\ref{fig:button_press}, Right). 
Although not statistically significant, we also observe that the autonomy has higher control authority (as measured by $\alpha$) during the disambiguation trials ($\alpha = 0.27 \pm 0.16$) when compared to the manual trials ($\alpha = 0.25 \pm 0.16$). In Equation~\ref{eq:sc}, larger $\alpha$ allocates more control authority to the autonomy. 

These results suggest that disambiguation is more useful as the control interface becomes more limited and the task becomes more complex. Intuitively, intent prediction becomes harder for the robot when the control interface is lower dimensional and does not reveal a great deal about the user's underlying intent. By having the users operate the robot in the disambiguating mode, the robot is able to elicit more intent-expressive control commands from the human which in turn helps in accurate goal inference and subsequently appropriate robot assistance. 

\noindent{\textbf{Temporal Distribution of Disambiguation Requests:}} 
\begin{table}[t]
	\centering
	\caption{Skewness of the temporal distribution of disambiguation requests. }
	\label{table:skewness}
	\begin{tabular}{|c|c|c|c|}
		\hline
		& Single-step & Multi-step \\
		\hline
		Joystick & 0.63 & 0.57 \\
		\hline
		Head Array & 0.35 & 0.22 \\
		\hline
	\end{tabular}
\end{table}
In Figure~\ref{fig:ha_man_disamb} the frequency and density of button presses (disambiguation requests plus mode switches) are much higher for the more limited control interface (head array). We observe that a higher number of disambiguation requests correlates with the more limited interface and complex task. The subjects also demonstrate a diverse range of disambiguation request behaviors, in regards to both (a) when the disambiguation requests are made and (b) with what frequency (e.g., Subject 1 vs. Subject 2, Joystick). The variation between subjects is likely due to different factors such as the user's comfort in operating the robot and understanding of the disambiguating mode's ability to recruit more assistance from the autonomy. 

The temporal distribution of disambiguation analyzes \textit{when} the subject requested assistance during the course of a trial. The skewness of the temporal distribution of disambiguation requests reveals a higher concentration of requests during the earlier parts of a trial (Table~\ref{table:skewness}) for both interfaces and tasks.\footnote{A uniform temporal distribution corresponds to a trial in which the disambiguation requests are uniformly spread out during the course of task execution. The skewness of a uniform distribution is zero.} However, under head array control the temporal distribution is less skewed, indicating that the need for disambiguation request persists throughout the trial, likely due to the extremely low-bandwidth of the interface.

\begin{figure}[t!]
	\centering
	\includegraphics[width = 1\hsize, ,center]{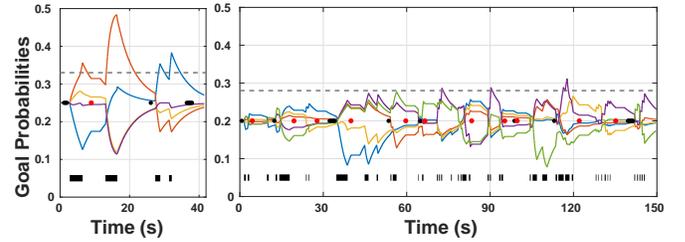}
	\caption{Time evolution of goal probabilities. Plot annotations include the minimum threshold for robot assistance (gray dashed line), disambiguation requests (red dots), manual mode switches (red dots), and indications of non-zero human control commands (black segments below the plotted data).}
	\label{fig:gp_evolution}
\end{figure}

\noindent{\textbf{Performance:}} 
\begin{figure}[b]
	\centering
	\includegraphics[width = 1\hsize, height=0.3\vsize, ,left]{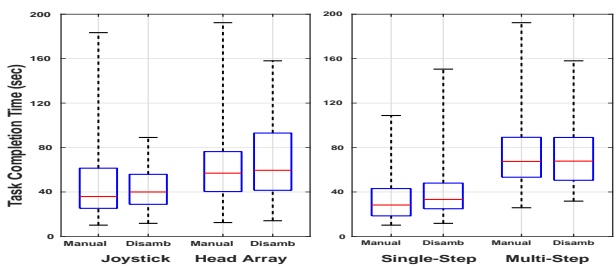}
	\caption{Task completion times. \textit{Disambiguation} and \textit{Manual} paradigms. \textit{Left:} Grouped by control interfaces. \textit{Right:} Grouped by tasks.}
	\label{fig:task_completion}
\end{figure}
No statistical difference is observed in task completion times between the \textit{Manual} and \textit{Disambiguation} paradigms (Figure~\ref{fig:task_completion}). However, the variance in the task completion times in general is lower under the \textit{Disambiguation} conditions (with the exception of single-step), indicating more consistent task performance when disambiguation requests are utilized. The task success is $92.50\%$ (148 out of 160 trials) and $97.92\%$ (94 out of 96 trials) for the disambiguation and manual paradigms respectively. 13 out of the 14 unsuccessful trials occur during the more complex multi-step task. Figure~\ref{fig:gp_evolution} provides illustrative examples of the time evolution of goal probabilities. Figure~\ref{fig:gp_evolution} (left) demonstrates how operation in the disambiguating mode can very quickly elevate one goal probability above the threshold for providing autonomy assistance. Figure~\ref{fig:gp_evolution} (right)
demonstrates how, at times, subjects do not leverage the capabilities of the disambiguating mode and immediately perform a manual mode switch, without ever issuing any control commands in the disambiguating mode.

\noindent{\textbf{User Survey:}} 
Table~\ref{table:survey_results} summarizes the results of the user survey. Users agree that task execution is easier during disambiguation trials (\textbf{Q1}, 4.88$\pm$0.95) and that operation under disambiguating modes is enjoyable (\textbf{Q3}, 5.00$\pm$1.15). User responses strongly validate the effectiveness of the blending-based shared control scheme (\textbf{Q2}, 6.19$\pm$0.75). Unsurprisingly, all users feel that it is harder to control the robot using the head array (\textbf{Q4}) and rate the utility value of the disambiguation paradigm to be higher for robot control with the head array (\textbf{Q5}). Although the subjects overwhelmingly prefer the \textit{Disambiguation} to the \textit{Manual} paradigm (\textbf{Q6}) only four out of the eight subjects find the \textit{Disambiguation} paradigm to be user-friendly (\textbf{Q7}). One possible explanation is a lack of transparency regarding why the autonomy chose the disambiguating mode.

\section{Discussion}\label{sec:discussion}
The disambiguation algorithm presented in our work can be utilized in any human-robot system in which there is a need to disambiguate between the different states of a discrete hidden variable (for example, a set of discrete goals in robotic manipulation or a set of landmarks in navigation tasks). Our algorithm assumes the existence of a discrete set of parameters (for example, control modes for robotic manipulation or natural language based queries for navigation) that can help the intent inference mechanism to precisely converge to the correct solution. Although the disambiguation algorithm is task-agnostic---because it relies exclusively on the shape features of the probability distribution over the hidden variable---the disambiguation is only as good as the efficacy of the inference algorithm that is used. In our experience, the choice of cost functions and domain-specific heuristics used for inference need to be appropriate for the task at hand. 
During our implementation development, the efficacy of the disambiguation algorithm degraded when we used only a subset of the four features to inform the disambiguation metric. This only reinforces the need for a combination of different shape features for successful disambiguation.

Another observation from our subject study is how often participants submitted a disambiguation request and then chose not to operate in the selected mode---effectively not letting the robot help them. This under-utilization phenomenon is illustrated in Figure~\ref{fig:gp_evolution} (right). One possible explanation is the subject's lack of understanding of how their control commands can help the robot to understand their intent. It is likely that a good grasp of the assistance mechanism is critical for providing intent-expressive control commands to the autonomy---underlining the need for extensive and thorough training, and for greater transparency in the human-robot interaction, so that the human has a clear understanding of how and why the autonomy chooses a specific assistance strategy.

The training can be made more effective in a few different ways.
For example, the subjects could be explicitly informed of the task relevant features (directedness, proximity \textit{et cetera}) that the autonomy relies on for determining the amount of assistance to offer. Knowledge of these features might motivate the users to leverage the disambiguating mode more.

The inherent time delays associated with the computation of the disambiguation mode (approximately 2-2.5s) might have been a cause for user frustration. Half of the subjects did report that the disambiguation system was not user-friendly. To improve upon this delay, a large set of disambiguating modes could be precomputed for different parts of the workspace, goal configurations and priors ahead of time, which then could be available in a lookup table during task execution. 
Automated mode switching schemes that eliminate the need for manual button presses altogether might also be a viable option for significantly reducing task effort. 

In our future work, as informed by our pilot study, we plan to extend the framework into an automated mode switch assistance system. A more extensive user study with motor-impaired subjects will also be conducted.

\begin{table}[t]
	\centering
	\caption{Subjective Survey Results}
	\label{table:survey_results}
	\begin{tabular}{|c|c|c|c|}
		\hline
		& Across Tasks & Single-step & Multi-step \\
		\hline
		\textbf{Q1} & 4.88 $\pm$ 0.95  & 4.88 $\pm$ 0.99 & 4.88 $\pm$ 0.99 \\
		\hline
		\textbf{Q2} & 6.19 $\pm$ 0.75 & 6.25 $\pm$ 0.89 & 6.13 $\pm$ 0.64 \\
		\hline
		\textbf{Q3} & 5.00 $\pm$ 1.15 & 5.25 $\pm$ 1.28 & 4.75 $\pm$ 1.03 \\
		\hline
		\textbf{Q4} & Head Array  & Head Array  & Head Array  \\
		\hline
		\textbf{Q5} & Head Array  & Head Array  & Head Array  \\
		\hline
		\textbf{Q6} & Disambiguation & Disambiguation & Disambiguation \\
		\hline
		\textbf{Q7} & Disamb/Manual & Disamb/Manual & Disamb/Manual\\
		\hline
	\end{tabular}
\end{table}
\section{Conclusion}\label{sec:conclusions}
In this paper, we have presented the idea of \textit{intent disambiguation assistance} via control mode selection for a shared-control assistive robotic arm. The aim of our control mode selection algorithm was to elicit more \textit{intent-expressive} control commands from the user by placing control in those control modes that \textit{maximally disambiguate} between the various goals in a scene. A pilot user study was conducted with eight subjects to evaluate the efficacy of the disambiguation system. Our results indicated a decrease in task effort in terms of the number of button presses when the disambiguation system was active. As our last contribution, we also presented a novel intent inference mechanism inspired by \textit{dynamic field theory} that works in conjunction with the disambiguation system.

\section*{Acknowledgment}
This material is based upon work supported by the National Science Foundation under Grant CNS 1544741. Any opinions, findings and conclusions or
recommendations expressed in this material are those of the authors and do
not necessarily reflect the views of the aforementioned institutions.

\balance
\bibliographystyle{IEEEtran}
\bibliography{GopinathArgall_TNSRE2020}

\end{document}